\def\year{2021}\relax
\documentclass[letterpaper]{article} 
\usepackage{aaai21}  
\usepackage{times}  
\usepackage{helvet} 
\usepackage{courier}  
\usepackage[hyphens]{url}  
\usepackage{graphicx} 
\usepackage{natbib}  
\usepackage{caption} 
\usepackage{bm}
\usepackage{booktabs}
\usepackage{multirow}
\usepackage[linesnumbered, ruled]{algorithm2e}
\usepackage{amsmath}
\usepackage[switch]{lineno}  %

\title{Finding Sparse Structures for Domain Specific Neural Machine Translation}
\author {
    Jianze Liang, \textsuperscript{\rm 1,2}\thanks{The work was done while JL was an intern at ByteDance AI Lab.}
    Chengqi Zhao, \textsuperscript{\rm 2}
    Mingxuan Wang, \textsuperscript{\rm 2} 
    Xipeng Qiu,  \textsuperscript{\rm 1} 
    Lei Li \textsuperscript{\rm 2}\\
}

\affiliations {
    \textsuperscript{\rm 1} School of Computer Science, Fudan University, Shanghai, China \\
    \textsuperscript{\rm 2} ByteDance AI Lab, China\\
    \{jzliang18, xpqiu\}@fudan.edu.cn,
    \{zhaochengqi.d, wangmingxuan.89, lileilab\}@bytedance.com
}

\newcommand{\method}{\textsc{Prune-Tune}\xspace}

\begin{document}

\maketitle

\begin{abstract}
Neural machine translation often adopts the fine-tuning approach to adapt to specific domains. 
However, nonrestricted fine-tuning can easily degrade on the general domain and over-fit to the target domain. 
To mitigate the issue, we propose \method, a novel domain adaptation method via gradual pruning. 
It learns tiny domain-specific sub-networks during fine-tuning on new domains. 
\method alleviates the over-fitting and the degradation problem without model modification.
Furthermore, \method is able to sequentially learn a single network with multiple disjoint domain-specific sub-networks for multiple domains. 
Empirical experiment results show that \method outperforms several strong competitors in the target domain test set without sacrificing the quality on the general domain in both single and multi-domain settings.
The source code and data are available at \url{https://github.com/ohlionel/Prune-Tune}.
\end{abstract}

\section{Introduction} \label{sec:intro}

Neural Machine Translation (NMT)  yields  state-of-the-art translation performance when a large number of parallel sentences are available~\cite{kalchbrenner2013recurrent, sutskever2014sequence, bahdanau2014neural, vaswani2017attention}. 
However, there are many language pairs lacking parallel corpora. 
It is also observed that NMT does not perform well in specific domains where the domain-specific corpora are limited, such as medical domain~\cite{koehn2007experiments,axelrod2011domain,freitag2016fast,chu2018survey}. 
There is huge need to produce high-quality  domain-specific machine translation  systems  whereas general purpose MT has limited performance.

Domain adaptation for NMT has been studied extensively. 
These work can be grouped into two categories: data-centric and model fine-tuning~\cite{chu2018survey}.  Data-centric methods focus on  selecting or generating target domain data from general domain corpora, which is effective and well explored~\cite{axelrod2011domain,chinea2017adapting,zeng2019iterative}. 
In this paper, we focus on the second thread: model fine-tuning.
Fine-tuning is very common in domain adaptation, which first trains a base model on the general domain data and then fine-tunes it on each target domain \cite{luong2015stanford,chen2017cost,gu2019improving,saunders2019domain}.
However, non-restricted  fine-tuning  requires very careful hyper-parameter tuning,  and is prone to over-fitting on the target domain as well as forgetting on the general domain. To tackle these issues, researchers have proposed several constructive approaches, with the view to limiting the size or plasticity of parameters in the fine-tuning stage, which can be roughly divided into two categories: regularization and partial-tuning strategy. Regularization methods often integrate extra training objectives to prevent parameters from large deviations, such as model output regularization \cite{khayrallah2018regularized}, elastic weight consolidation (EWC) \cite{thompson2019overcoming}. Regularization methods, which impose arbitrary global constraints on parameter updates, may further restrict the adaptive process of the network, especially when domain-specific corpora are scarce. Partial-tuning methods either freeze several sub-layers of the network and fine-tune the others~\cite{thompson2018freezing}, or integrate domain-specific adapters into the network~\cite{bapna2019simple, vilar2018learning}. By only fine-tuning the domain-specific part of the model, they can alleviate the over-fitting and forgetting problem in fine-tuning. However, the structure designed to adapting is usually hand-crafted, which relies on experienced experts and the adapter brings additional parameters. Therefore,  a more adaptive, scalable, and parameter-efficient approach for domain adaptation is very valuable and worth well studying.

\begin{figure*}[!ht]
        \centering
        \includegraphics[width=0.9\linewidth]{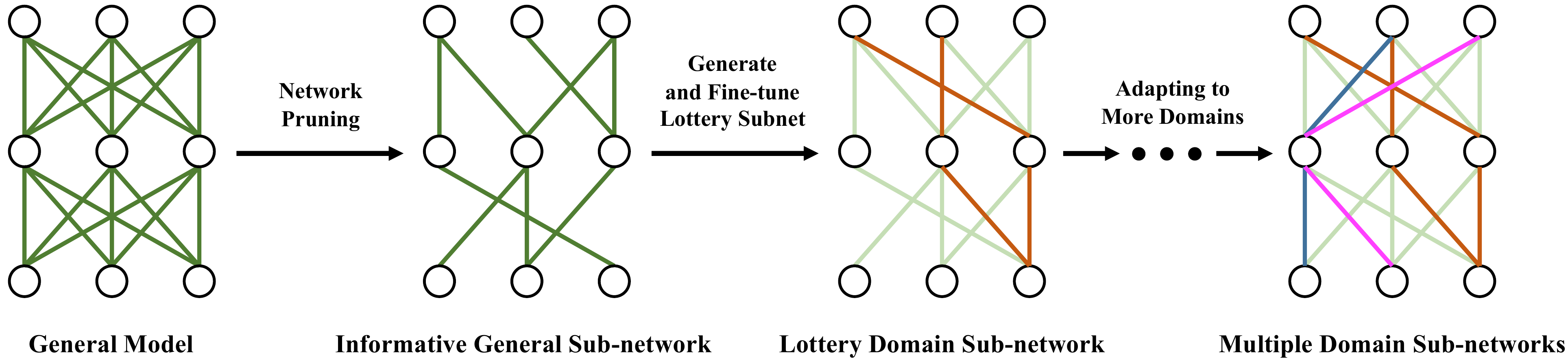}
        \caption{Illustration of domain adaptation from the general domain to the target domain with \method. c)$\rightarrow$ d) demonstrates our proposed \method is capable of adapting to multiple domains.} 
        \label{Fig:method}
    \end{figure*}

In this paper, we propose \method, a novel domain adaptation method via adaptive structure pruning. Our motivation is inspired from Continual Learning \cite{parisi2019continual, kirkpatrick2017overcoming, mallya2018packnet, mallya2018piggyback, hung2019compacting, lee2019mixout} and \emph{the lottery hypothesis} that a randomly-initialized, dense neural network contains a sub-network which  can match the test accuracy of the original network after training for at most the same number of iterations~\cite{frankle2018lottery}.  
We therefore suppose that multiple  machine translation models for different domains can share different sparse sub-networks within a single neural network.  
Specifically, we first apply a standard pruning technique to automatically uncover the sub-network from a well-trained NMT model in the general domain.  The  sub-network is capable of  reducing the parameter without compromising accuracy. Therefore, it has the potential to keep as much general information as possible.  
Then we freeze this informative sparse network and leave the unnecessary  parameters unfixed for the target  domain, which enables our approach to be parameter efficient, and eases the scalability of the approach to more domains. 
The capacity of these non-fixed parameters can be tuned to match the requirements of the target domain, while keeping the parameters of the general domain. Our method successfully circumvents catastrophic forgetting problem~\cite{kirkpatrick2017overcoming} and retains the quality on the general domain. 
As the benefits of the flexible design, \method can be easily extended to other transfer learning problems, such as multilingual machine translation.   

We summarize our main contribution as follows:
\begin{itemize}
    \item We propose \method,  which enables generating domain-specific sub-networks via gradual pruning  and potentially  circumvents  the notorious \emph{catastrophic forgetting} problems in domain adaptation.
    \item We conduct extensive experiments to  evaluate \method  and demonstrate that \method outperforms the strong competitors both in the general and target domain with big margins. On domain adaptation benchmarks for \textsc{En$\rightarrow$De}, \method outperforms several strong competitors including Fine-tuning, EWC, Model Distillation, Layer Freeze and Adapter in target domain test set without the loss of general domain performance.
    \item We extend \method to multi-domain experiments on \textsc{En$\rightarrow$De} and \textsc{Zh$\rightarrow$En}, which shows the possibilities of training a single model to serve different domains without performance degradation.   
    
\end{itemize}

\section{Background}
\subsection{Neural Machine Translation}
Given an source sentence $\mathbf{x} = \{x_1, \dots , x_n\}$ and its translation $\mathbf{y}=\{y_1, \dots , y_m\}$, Neural Machine Translation directly models the conditional probability of target sentence over source sentence:
\begin{equation}
    P(\mathbf{y}|\mathbf{x};\theta) = \prod_{i=1}^m P(y_i|\mathbf{x}, y_{<i}; \theta),
\end{equation}
where $\theta$ denotes the parameters of the model.
For a parallel training dataset $D={\{\mathbf{x}^j,\mathbf{y}^j\}}_{j=1}^N$, $\theta$ is optimized to maximum the log-likelihood:
\begin{equation}
    \mathop{\arg\max}_{\theta} \sum_{j=1}^N \log P(\mathbf{y}^j|\mathbf{x}^j;\theta).
\end{equation}

\subsection{Fine-tuning for Domain Adaptation} 
Model fine-tuning on the target domain is the most natural approach for domain adaptation. Assume we have a well trained NMT model $\mathcal{F}(\cdot;\theta)$ and a dataset $D_{I}={\{\mathbf{x}^i,\mathbf{y}^i\}}_{i=1}^{N_I}$ of a new domain. We can simply apply fine-tuning to adapt the model to the new domain, that is, we continue training the model to optimize $\theta$ on $D_{I}$:
\begin{equation}
    \mathop{\arg\max}_{\theta} \sum_{i=1}^{N_I} \log P(\mathbf{y}^i|\mathbf{x}^i;\theta).
\end{equation}
As discussed in Introduction, fine-tuning on all model parameters $\theta$ often leads to over-fitting on the new domain as well as forgetting on the general domain. So apart from regularization approaches, it is effective to introduce domain-specific part of models to alleviate these problems. There are two typical kinds of methods: layer freeze and adapter.

\textbf{Layer freeze} approaches regard the top layer, denoted as $\theta_{L}$, of model as the domain-specific parameters while the rest parameters $\theta_{l<L}$ are kept fixed. The training object of layer freeze is:
\begin{equation}
\label{layer_freeze_eqa}
    \mathop{\arg\max}_{\theta_{L}} \sum_{i=1}^{N_I} \log P(\mathbf{y}^i|\bm{H}_{L-1};\theta_{L}),
\end{equation}
where $\bm{H}_{L-1} = \mathcal{F}(\mathbf{x}^i; \theta_{l<L})$ indicates the output of the $\left(L-1\right)$-th layer of the model.

\textbf{Adapter} methods integrate an additional module $\theta_A$ into the network. The additional module can be a fully-connection layer, a self-attention layer or their combinations. Finally we fine-tune only on the domain-specific part $\theta_A$ and the training objective is as follow:
\begin{equation}
\label{adapter_equ}
    \mathop{\arg\max}_{\theta_{A}} \sum_{i=1}^{N_I} \log P(\mathbf{y}^i|\bm{H}_{L};\theta_{A}),
\end{equation}
where $\bm{H}_{L} = \mathcal{F}(\mathbf{x}^i; \theta)$.

As shown in equation \eqref{layer_freeze_eqa} and \eqref{adapter_equ}, domain-specific parameters only interact with the output of general model, i.e. $\mathcal{F}(\cdot;\theta)$. We suppose interaction more with the general model would achieve much better performance.

\section{Approach}

As many studies show, a great proportion of parameters in the network are redundant \cite{frankle2018lottery, zhu2017prune, liu2018rethinking}. 
Pruning such parameters causes minor or even no degradation in the task.
\citet{zhu2017prune} show that the dynamic and sparse sub-network after pruning is expressive and outperforms the dense network with the equivalent size of parameters.
Therefore, it is possible to make use of such redundancy for domain adaptation.

Given a well trained general model, our approach consists of the following  steps (see Figure \ref{Fig:method}): 
\begin{enumerate}
    \item Find and freeze   the most informative parameters of the general  domain and leave unnecessary parameters for the target domain
    \item Uncover the lottery sub-networks from the free parameters  for a specific domain
    \item Tune the lottery sub-networks for the specific domain
    \item Repeat the 2-3 steps  for multi-domain adaptation
\end{enumerate}


\subsection{Finding the Informative Parameters for General Domain}
\begin{figure}[t]
        \centering
        \includegraphics[width=1.0\linewidth]{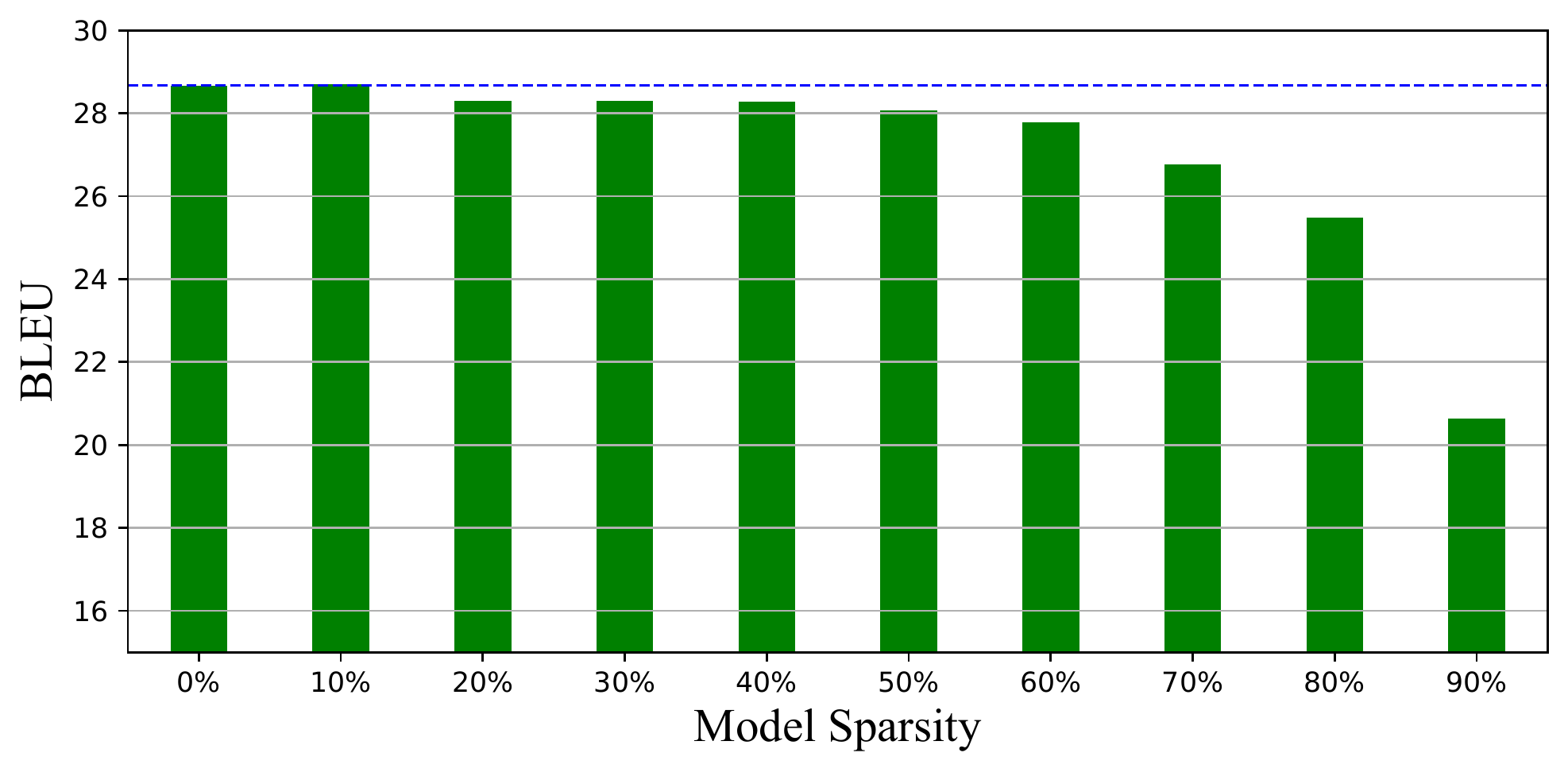}
        \caption{BLEU scores of pruned Transformer models with different sparsity (percentage of pruned parameters) on WMT14 \textsc{En$\rightarrow$De}.  Notice that even with 50\% of the original parameters, the resulting model still achieves nearly the same translation performance as the original full Transformer. }
        \label{Fig:pruning}
    \end{figure}

Pruning has proven to be effective  for  keeping the informative parameters and eliminating unnecessary ones for neural networks~\cite{lecun1990optimal,li2016pruning,han2015learning,zhu2017prune}.
Without loss of generality, we employ a simple and effective \emph{Gradual Pruning}  approach to find the most informative  parameters for the general domain~\cite{zhu2017prune}.
The method gradually prunes the model to reach the target sparsity by reducing low magnitude parameters every 100 training steps.
Explicitly, we trim parameters to the target sparsity in each layer. 
Between pruning steps, the model is trained on the general dataset to recover its performance in the sub-network.
Though NMT is one of the most complicated tasks in deep learning, our empirical study on pruning sparsity shows that up to 50\% parameters in a Transformer big model are not necessary and can be pruned with a performance drop less than 0.6 BLEU (see Figure \ref{Fig:pruning}).
In this way, we can keep the general NMT model intact as an informative sub-network  of the original  model.
To keep consistent generalization ability provided from the original  sub-network, we freeze the parameters of the informative  sub-network during  domain adaptation process.

The left unnecessary  weights throughout the network provide the possibility  of generating a sparse lottery  sub-network that can exactly  match the test accuracy of the domain-specific model.
As the lottery sub-network keeps most of the general domain information, fine-tuning the unnecessary weights can potentially outperform the full fine-tuning approach. 
Particularly, the sparsity rate is very flexible which can be changed to meet the requirements  of various scenarios. 
In general, a low sparsity rate is suitable for simple domain adaptation tasks, while high sparsity works better for  complicated domain or multiple domain adaptation tasks.

\subsection{Lottery Sub-network Generation for Specific Domain}
It is not necessarily needed  to fine-tune all the free parameters for a specific domain, especially for multi-domain adaptation tasks that require parameter efficient sub-networks for different domains. 
As the extracted informative sub-network already has a strong capacity, we suppose that a few additional parameters may be enough for the target domain adaptation. 
The most challenging problem is to automatically  uncover the best sparse structure  for the specific domain within. And we call this sparse structure as \emph{lottery sub-network}.   
The challenge  is essentially a network architecture search problem (NAS) to learn domain-specific sub-network, which is very costly. 
For simplicity, we  apply an iterative pruning method again as an effective way to learn the  lottery sub-network. 

Specifically, 
we fine-tune the free parameters on the target domain data for a few steps as warm-up training, then apply pruning to obtain the domain-specific structure.  The generated structure is then fixed as the lottery domain-specific  sub-network for further fine-tuning.   

\subsection{Fine-tuning of Domain-Specific Sub-network}
We introduce a mask matrix over all parameters in the network which indicates the sub-network for each domain with different domain identification.
Each parameter of the network belongs to only one specific domain, and can not be updated by learning of other domains.  

For single domain adaptation, we adapt the general domain to the target domain by training on the combined parameters of the general informative sub-network  and the domain-specific  lottery sub-network.
For multiple domain adaptation, we iteratively repeat this process based on the general model. 
It is rather flexible as we do not require data from all domains simultaneously. 
Particularly, with the partition of parameters, we can adapt a new domain only from helpful domains. 
Supposes that we have successfully trained a multi-domain system supporting three different domains: news, law, biology. 
While our goal is to adapt to a new medical domain, it is capable of  incorporating both the general  and biology domain as source domains for the medical domain. 

\method shares different domain sub-network in a single transform model with domain-specific masks. 
Given the source sentence and the corresponding  domain identification, a binary domain mask will be applied to the unified model to support decoding with only the learned sparse sub-network. The mask matrix makes the system rather flexible for practical application or extends to a new domain.

\section{Experiment}
We conducted experiments on both single domain adaptation and multiple domain adaptation to show the effectiveness and flexibility  of \method. 

\subsection{Dataset}

To evaluate our model in single domain adaptation, we conducted experiments on English to German translation, where the training corpora for the general domain were from WMT14 news translation task. And we used newstest2013 and newstest2014 as our validation and test set respectively. 
The general domain model trained on \textsc{WMT14 En$\rightarrow$De} was then individually adapted to three distinct target domains: TED talks, biomedicine, and novel. 
For TED talks, we used IWSLT14 as training corpus, dev2010, and tst2014 as the validation and test set respectively. 
For the biomedicine domain, we evaluated on EMEA News Crawl dataset\footnote{\url{https://ufal.mff.cuni.cz/ufal_medical_corpus}}. 
As there were no official validation and test set for EMEA, we used Khresmoi Medical Summary Translation Test Data 2.0\footnote{\url{https://lindat.mff.cuni.cz/repository/xmlui/handle/11234/1-2122}}.
For novel domain, we used a book dataset from OPUS\footnote{\url{http://opus.nlpl.eu/}} \cite{tiedemannparallel}. We randomly selected several chapters from \textit{Jane Eyre} as our validation set and \textit{The Metamorphosis} as the test set.

We extended \method to multi-domain adaptation on English to German and Chinese to English translation. For \textsc{Zh$\rightarrow$En}, we used the training corpora from \textsc{WMT19 Zh$\rightarrow$En} translation task as the general domain data. We selected 6 target domain datasets from from UM-Corpus\footnote{\url{http://nlp2ct.cis.umac.mo/um-corpus}} \cite{tian2014corpus}.

Table \ref{tb:data} lists the statistics of all datasets mentioned above.

\begin{table}[!t]\small
  \centering
  \begin{tabular}{ccccc}
    \toprule
    Direction & Corpus & \multicolumn{1}{c}{Train} &  \multicolumn{1}{c}{Dev.} & \multicolumn{1}{c}{Test}   \\
    \midrule
    \multirow{4}*{\textsc{En$\rightarrow$De}}    & WMT14         & 3.9M   & 3000  & 3003  \\
                                        & IWSLT14       & 170k  & 6750	& 1305  \\
                                        & EMEA	        & 587k  & 500   & 1000  \\
                                        & Novel	        & 50k   & 1015  & 1031  \\
    \midrule
    \multirow{7}*{\textsc{Zh$\rightarrow$En}}    & WMT19	        & 20M  & 3000   & 3981    \\
                                        & Laws	        & 220k  & 800   & 456  \\
                                        & Thesis	    & 300k  & 800	& 625   \\
                                        & Subtitles     & 300k  & 800   & 598      \\
                                        & Education     & 449K  & 800   & 791      \\
                                        & News          & 449K  & 800   & 1500      \\
                                        & Spoken        & 219k  & 800   & 456      \\
    \bottomrule
  \end{tabular}\caption{Datasets statistic for En$\rightarrow$De and Zh$\rightarrow$En tasks.}\label{tb:data}
\end{table}

\subsection{Setup}
For \textsc{En$\rightarrow$De} data preprocessing, we tokenized data using \textit{sentencepiece}~\cite{kudo2018sentencepiece}, with a jointly learned vocabulary of size 32,768. For \textsc{Zh$\rightarrow$En}, we applied jieba and moses tokenizer to Chinese and English side respectively. Then we encoded sentences using byte pair encoding (BPE)~\cite{sennrich2016neural} with 32k merge operations separately. We implemented our models on recently the state-of-the-are translation model, Transformer~\cite{vaswani2017attention} and we followed the big setting, including 6 layers for both encoder and decoders. The embedding dimension was 1,024 and the size of ffn hidden units was 4,096. The attention head was set to 16 for both self-attention and cross-attention. We used Adam optimizer~\cite{kingma2014adam} with the same schedule algorithm as \citet{vaswani2017attention}. All models were trained with a global batch size of 32,768 on NVIDIA Tesla V100 GPUs. During inference, we used a beam width of 4 for both \textsc{En$\rightarrow$De} and \textsc{Zh$\rightarrow$En} and we set the length penalty to 0.6 for \textsc{En$\rightarrow$De}, 1.0 for \textsc{Zh$\rightarrow$En}.

The evaluation metric for all our experiments is tokenized BLEU \cite{papineni2002bleu} using \textit{multi-bleu.perl}\footnote{\url{https://github.com/moses-smt/mosesdecoder/blob/master/scripts/generic/multi-bleu.perl}}. 

\begin{table*}[!ht]\small
  \centering
  \begin{tabular}{lcccccccc}
    \toprule
    \multirow{2}*{Model} & \multicolumn{2}{c}{IWSLT (190k)}  &  \multicolumn{2}{c}{EMEA (587k)} &  \multicolumn{2}{c}{Novel (50k)} & \multirow{2}{*}{\#Tuning Params} \\
     \cmidrule(lr){2-3} \cmidrule(lr){4-5} \cmidrule(lr){6-7}  
         & general   & target & general   & target & general  & target &    \\
  \midrule
    Target Domain Model      & 11.4   & 24.0   & 3.1  & 23.9 & 2.7   & 12.3 & 273M      \\
  \midrule
    Mixed Domain Model      & 27.9  & 31.3  & 27.9   & \textbf{32.0} & 27.9   & 21.2 & 273M      \\
    
  \midrule
    General Domain Model     & 28.7 & 28.5 & 28.7 & 28.4 & 28.7 & 14.5 & 273M    \\
    + Fine-tuning \cite{luong2015stanford} & 27.0 & 31.5 & 17.1 & 29.7 & 12.1 & 23.4 & 273M   \\
    + EWC-regularized~\cite{thompson2019overcoming}                  & 28.0 & 31.5 & 27.1 & 30.5 & 23.5 & 23.1 & 273M   \\
    + Model Distillation~\cite{khayrallah2018regularized} & 26.3 & 31.5 & 16.3 & 30.0 & 11.6 & 23.1 & 273M \\
    + Layer Freeze~\cite{thompson2018freezing}          & 28.6 & 31.3 & 26.9 & 29.8 & 23.0 & 23.0 & 29M   \\
    + Adapter~\cite{bapna2019simple}               & 27.0 & 31.6 & 26.7 & 30.1 & 19.8 & \textbf{24.3} & 13M  \\
    
    \midrule
    \method Model                   &  \textbf{28.8} & \textbf{31.9}  & \textbf{28.9}  & 30.6  & \textbf{28.8}  &  \textbf{24.3} & 27M   \\
    \bottomrule
  \end{tabular}\caption{BLEU scores of single domain adaptation on \textsc{En$\rightarrow$De}. All models share the same Transformer-big setting. Notice that \method improves the translation performance on the specific domains while maintaining the general domain performance.   }\label{tab:single_result}
\end{table*}

\subsection{Domain Adaptation on Single Lottery Sub-network}

We used a lottery sub-network with 10\% sparsity and conducted domain adaptation experiments on \textsc{En$\rightarrow$De}. The 10\% free parameters were tuned to fit each target domain. During inference, our model can recover the capability of the general domain by simply masking these domain-specific parameters. We compared our model with several strong baselines and effective models:

\begin{itemize}
    \item \textbf{{General domain model}}: The model was trained using only parallel data from the general domain.
    \item \textbf{{Target domain model}}: The model was trained using only the target domain data.
    \item \textbf{{Mixed domain model}}: All general domain and target domain data were mixed to train the model.
    \item \textbf{{Fine-tuning}} \cite{luong2015stanford}: We continued to train the general domain model on target domain data with the training step unchanged. The empirical study shows it performs better than resetting the training step to 0.
    \item \textbf{{EWC-regularized model}} \cite{thompson2019overcoming}: EWC \cite{kirkpatrick2017overcoming} is a popular algorithm in Continual Learning \cite{parisi2019continual}, which applies elastic consolidation to each parameter during gradient updates. The EWC-regularized model prevents the parameters from large deviations.
    \item \textbf{Model Distillation} \cite{khayrallah2018regularized}: We employed an auxiliary loss during fine-tuning to prevent the target domain model's output from differing too much from the original general domain model's output.
    \item \textbf{{Layer Freeze}} \cite{thompson2018freezing}: We froze all model layers except for the top layers of both the encoder and decoder, then fine-tuned the top layers on the target domain data. 
    \item \textbf{{Adapter}} \cite{bapna2019simple}: We stacked adapters on each transformer block of both encoder and decoder as proposed by \citet{bapna2019simple}, and fine-tuned the adapters only.
\end{itemize}

\begin{figure}[t]
        \centering
        \includegraphics[width=1.0\linewidth]{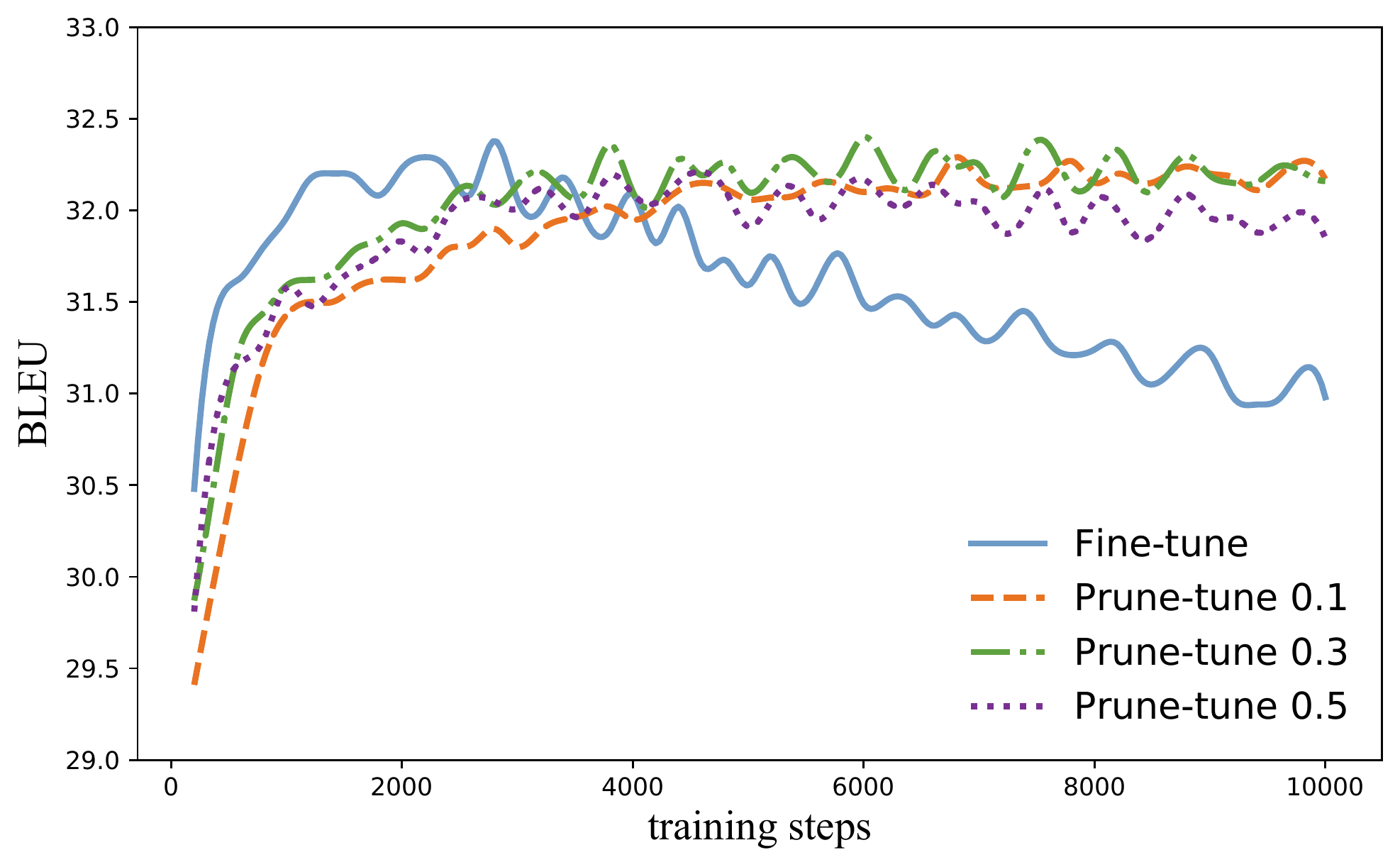}
        \caption{BLEU scores of fine-tuning and our proposed \method with 10\%, 30\%, 50\% sparsity when adapting to IWSLT14 \textsc{En$\rightarrow$De}. Notice the plain fine-tuning will degrade in the end on the target domain, while \method steadily improves.}
        \label{Fig:performance}
    \end{figure}

Our proposed \method outperforms fine-tuning and other baselines as shown in Table \ref{tab:single_result}.
In three distinct domains with varying corpus size, our approach achieves competitive performance with less training parameters.
Moreover, our model is able to serve both general or target domain machine translation without any performance compromise in a unified model, simply via a domain mask. 
Figure \ref{Fig:performance} demonstrates that our approach effectively alleviates the serious over-fitting problem that fine-tuning often suffers from. To conclude, \method enjoys the following advantages:
\begin{itemize}
    \item \method is very effective for the target domain adaptation.  We attribute this to the adaptive pruning  of the  lottery sub-network.  With little modification of a  sub-network, \method significantly  outperforms \emph{Layer Freeze} and \emph{adapter} with pre-defined sub-network fine-tuning, which shows the benefits of dynamic structure finding. 
    \item Clearly, \method is firmly capable of  keeping the translation performance in the general domain. After fine-tuning on  the novel domain, \method even surpasses  the second fine-tuning competitor by 5 BLEU score in the general domain. 
    \item \emph \method ~is robust when compared to the fine-tuning baseline, which suffers from the over-fitting challenges and requires very careful checkpoints choices.  
\end{itemize}

\begin{table*}[!ht]
  \centering
  \begin{tabular}{lclcccccc}
    \toprule
       Model  & \multicolumn{2}{c}{Input domain} & \#M  & WMT14 (\textbf{W}) & IWSLT (\textbf{I})  & EMEA (\textbf{E})  & Novel (\textbf{N})   \\
    \midrule
        Mixed Domain Model
        & \multicolumn{2}{c}{\textbf{W}, \textbf{I}, \textbf{E}, \textbf{N}}   & 1   & 27.9  & 31.3  & 32.0  & 21.2    \\
    \midrule
        General Domain Model
        & \multicolumn{2}{c}{\bf{W}}     &  1   & 28.7  & 28.5  & 28.4  & 14.5    \\
        
        + Fine-tuning
        & \multicolumn{2}{c}{\bf{I}, \bf{E}, \bf{N}}   & 3  & N/A  & 31.5  & 29.7  & 23.4    \\
    \midrule
        Single \method Model
        & \multicolumn{2}{c}{\bf{W}, \bf{I}, \bf{E}, \bf{N}}   & 3     & N/A  & 31.9  & 30.6  & 24.3    \\
    \midrule
        \multirow{4}*{Sequential \method Model}  
        & \#1   & \bf{W}        & \multirow{4}*{1}   & 28.4  & N/A  & N/A  & N/A    \\
        & \#2   & + \bf{I}   &    & 28.4  & 31.9 & N/A  & N/A      \\
        & \#3   & + \bf{E}    &   & 28.4  & 31.9 & 30.1 & N/A     \\
        & \#4   & + \bf{N}   &   & 28.4  & 31.9 & 30.1 & 23.6  \\
    
    \bottomrule
  \end{tabular}\caption{BLEU scores of sequential domain adaptation on \textsc{En$\rightarrow$De}. \#M denotes the number of required models. {W}, {I}, {E}, {N} refer to dataset WMT14, IWSLT, EMEA, Novel, respectively. In our sequential \method  Model, general domain occupied 50\% parameters, and each target domain occupied 10\%. Notice that sequential \method obtains a single model with best performance on all domains except EMEA.}\label{tab:multi_result_en2de}
\end{table*}

\begin{table*}[!ht]
  \centering
  \begin{tabular}{lccccccccc}
    \toprule
        Model & \#M & Laws  & Thesis  & Subtitles & Education & News & Spoken & Avg.   \\
    \midrule
        Mixed Domain Model      & 1  & 47.4 & 15.6 & 17   & 31.4 & 21.2 & 16.7 & 24.9  \\
    \midrule
        General Domain Model    & 1  & 44.9 & 13.8 & 16.1 & 30.8 & 21.4 & 16.7 & 23.9  \\
        + Fine-tuning             & 6  & 55.9 & 17.9 & 20.8 & 29.2 & 22.1 & 14.8 & 26.7      \\
        \midrule
        Sequential \method Model     & 1  & 50.3 & 16.2 & 17.2 & 31.2 & 21.3 & 14.6 & 25.1 \\
    
    \bottomrule
  \end{tabular}\caption{BLEU scores of sequential domain adaptation on \textsc{Zh$\rightarrow$En}. \#M denotes the number of required models. In our Sequential \method Model, general domain occupied 50\% parameters, and each target domain occupied 5\%. Notice that Sequential \method is the best performing single model for all domains.  }\label{tab:multi_result_zh2en}
\end{table*}

\subsection{Sequential Domain Adaptation}
We conducted multi-domain adaptation experiments on \textsc{En$\rightarrow$De} and \textsc{Zh$\rightarrow$En} to demonstrate the unique sequential learning ability of our approach.

We first trained general models on \textsc{En$\rightarrow$De} and \textsc{Zh$\rightarrow$En}, and then gradually pruned them to reach 50\% sparsity. 
We find it empirically that 50\% is a sparsity with no significant performance drop and enough redundant parameters.
Different from single domain adaptation, we fixed embedding layers and layer normalization parameters to avoid sharing parameters across multiple domains.
In these experiments, we adopt the general models to target domains sequentially.
For each target domain:
\begin{enumerate}
    \item Firstly, we applied warm-up training and \emph{Gradual Pruning} to generate a suitable lottery domain sub-network.
    \item Secondly, We simply adapt the general domain learned before to the current domain by including the general sub-network as frozen parameters.
    \item Finally, we fine-tune the lottery sub-network of the domain.
\end{enumerate}

We adapted 3 target domains on \textsc{En$\rightarrow$De}, and 6 target domains on \textsc{Zh$\rightarrow$En}.

\subsubsection{Result}
In Table \ref{tab:multi_result_en2de}, we report performance in \textsc{En$\rightarrow$De} experiment. 
Within a single model, our approach can learn new domains in sequence and outperforms several baselines.
Specifically, it outperforms Mixed Domain Model which requires all domain data simultaneously, and Fine-tuning which requires multiple models.
Moreover, \method can learn the current target domain while retaining the performance in previously learned domains, because lottery domain sub-networks are separate.

\textsc{Zh$\rightarrow$En} experiment result in Table \ref{tab:multi_result_zh2en} also demonstrates that our approach is effective and flexible for more domains.
 

\section{Analysis}
In this section, we revisit our approach to reveal more details and explain the effectiveness of the proposed \method.

\subsection{Robustness of \method}
We are convinced that the over-fitting problem seriously affects the robustness of fine-tuning.
As shown in figure \ref{Fig:performance}, fine-tuning reaches the best performance at the early step, and then starts to decline, while our method yields stable performance.
When the target data is scarce, domain adaptation by unrestricted fine-tuning will rapidly over-fit to the target domain, forgetting the generalization ability from the general model.
Our proposed \method is a more robust method as we integrate a frozen informative sub-network within the model, which provides generalized information consistently.


\begin{table}[t]
  \centering
  \begin{tabular}{ccccc}
    \toprule
    \multicolumn{1}{c}{Pruning Rate}& \multicolumn{1}{c}{WMT} & \multicolumn{1}{c}{IWSLT} &  \multicolumn{1}{c}{EMEA} & \multicolumn{1}{c}{Novel}   \\
    \midrule
        10\%    & \textbf{28.7} & 32.3 & \textbf{30.6} & \textbf{24.3}  \\
        30\%    & 28.3 & \textbf{32.4} & 30.3	& 23.9  \\
        50\%	& 28.1 & 32.2 & 29.5 & 23.6  \\
        70\%	& 26.8 & 31.8 & 28.9 & 23.1 \\
    \bottomrule
  \end{tabular}\caption{BLEU scores of different pruning rate for \method. Only 10\% of parameters for fine-tuning is able to achieve the best performance.}\label{tb:sparsity}
\end{table}

\subsection{Less Pruning Improves Performance}
Since we can prune the model to different sparsity, we evaluate the single domain adaptation performance on general models with different sparsity. 
As shown in Table \ref{tb:sparsity}, domain adaptation on low sparsity  achieves better performance mainly due to better knowledge preservation of the general domain. 
It also indicates that a few parameters are enough for single domain adaptation. 
As the pruning goes further, the high sparsity model is doomed to degrade on the general domain, which affects the subsequent domain adaptation. 
However, the performance gap between low sparsity \method models is relatively small.

\subsection{\method is Very Effective for Low-resource Domain Adaptation }
\begin{figure}[t]
        \centering
        \includegraphics[width=1.0\linewidth]{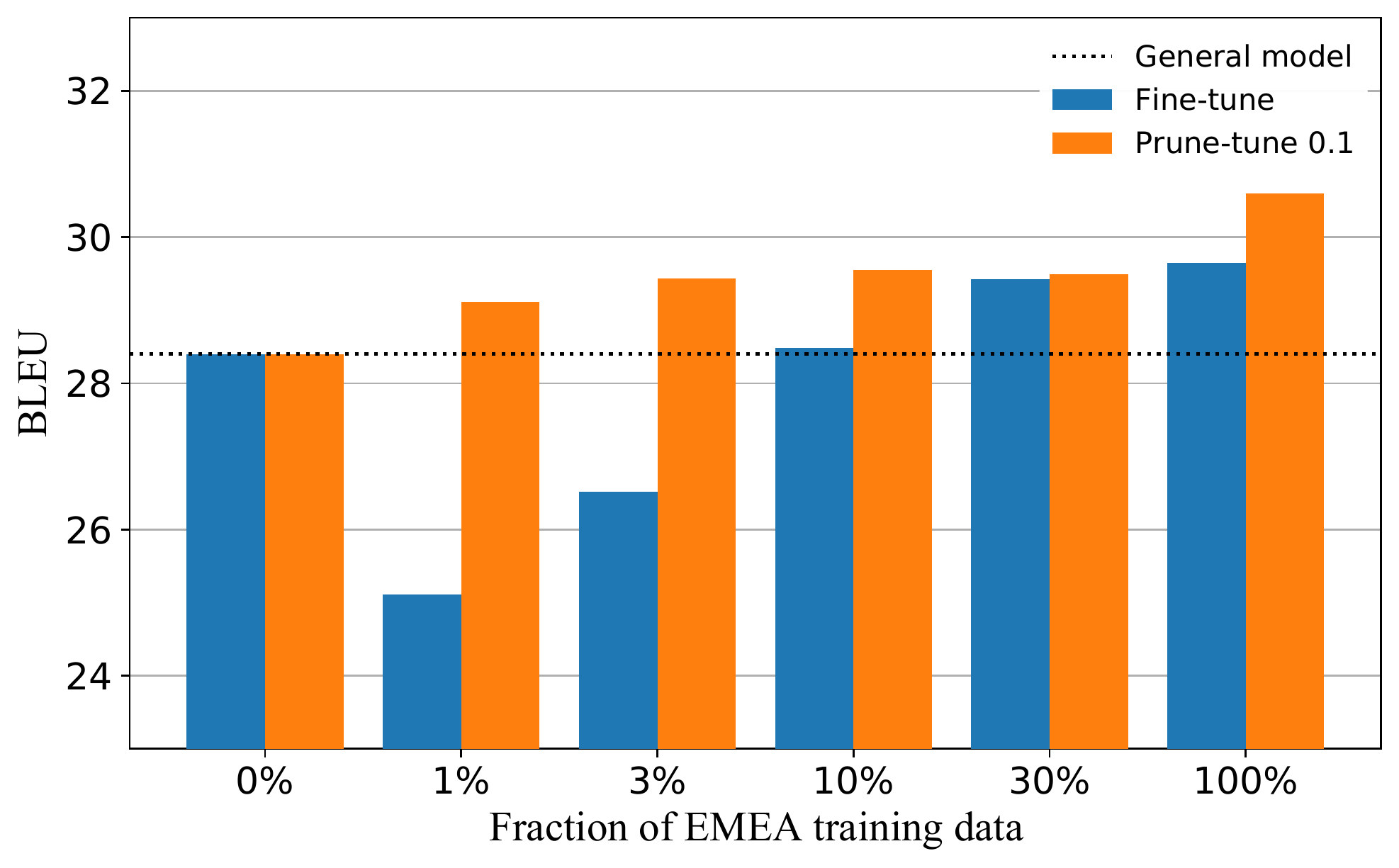}
        \caption{Fine-tuning  with different domain-specific corpus.   \method improves the baseline at different scales, while full fine-tuning suffers from over-fitting. }
        \label{Fig:data-size}
    \end{figure}

To evaluate the performance of our approach on varying amounts of target domain data, we experimented on the EMEA dataset with different fractions of training data.
We extract 1\%, 3\%, 10\%, 30\% and 100\% of the original EMEA training set.
We compare with full fine-tuning using 10\% sparsity \method model on different fractions of EMEA dataset.
As the results are shown in Figure \ref{Fig:data-size}, our approach significantly outperforms fine-tuning for each fraction.
Especially for extremely small 1\% fraction, which consists of 5.7K sentences, our proposed approach improves the performance over the general model by 0.7 BLEU, while fine-tuning leads to a 3.3 BLEU drop.
With fractions less than 30\%, fine-tuning can not improve the target bio domain, but brings damage to the general model.
In the contrast, our approach does not harm the general domain ability, and can make the most of the few training data to improve the target domain.  
It indicates that our proposed approach is suitable for low resource domain adaptation, which is common and valuable in practice.

\begin{table}[t]
  \centering
  \begin{tabular}{cc}
    \toprule
    Adaptation Order  & EMEA    \\
    \midrule
        1   & 30.3   \\
        2   & 30.1   \\
        3	& 30     \\
    \bottomrule
  \end{tabular}\caption{BLEU scores of different adaptation order for sequential domain adaptation.}\label{tb:domain order}
\end{table}

\begin{table}[t]
  \centering
  \begin{tabular}{ccccc}
    \toprule
    Target Domain Params(\%) & IWSLT & EMEA & Novel   \\
    \midrule
        1\%     & 31.7 & 29.4 & 22.3  \\
        5\%	    & 31.8 & 30.1 & 23   \\
        10\%	& 31.9 & 30.1 & 23.6  \\
    \bottomrule
  \end{tabular}\caption{BLEU scores of different scale of target domain-specific parameters for sequential domain adaptation.}\label{tb:domain params}
\end{table}

\subsection{Sequential \method is Capable for Numerous Domains} 
We conducted experiments to explore the limit of sequential multi-domain adaptation with \method. We first evaluated the influence of the learning order of the EMEA dataset. As shown in Table \ref{tb:domain order}, there is only a minor gap of BLEU score between different learning order. We also conducted experiment on \textsc{En$\rightarrow$De} with different scale of target domain-specific parameters. As shown in Table \ref{tb:domain params}, 5\% of parameters is sufficient for most domains, and even 1\% of parameters yields comparable performance. Actually, \method has the potential to adapt to dozens of domains. 

\section{Related Work}

\subsection{Domain Adaptation}
Domain adaptation has been widely investigated in recent years. 
In Machine Translation, the fine-tuning based approach is the most relevant to our work. 
Fine-tune is the conventional way for domain adaptation \cite{luong2015stanford, sennrich2016improving, freitag2016fast, chu2017empirical}.
Many studies try to address the shortcoming of Fine-tune.
\citet{thompson2018freezing} freeze selected modules of the general network.
Adapters is introduced for parameter efficiency \cite{bapna2019simple, vilar2018learning}.
\citet{khayrallah2018regularized} explore regularization techniques to avoid over-fitting.
\citet{thompson2019overcoming} employ EWC \cite{kirkpatrick2017overcoming} to alleviate the catastrophic forgetting problem in domain adaptation. \citet{zhang2020revisiting} re-initialize parameters from some layer for few-sample BERT fine-tuning.
\citet{wuebker2018compact} introduce sparse offset from the general model parameters for every domain, sharing the similar idea of our proposed method. The key difference is that \method provides a dynamic parameter adaptation method, which is  parameter efficient and potentially makes the most of general domain information for the target  domain. 

Another research line for domain adaptation is data selection and data mixing, both being concerned with how to sample examples to train an MT model with a strong focus on a specific domain~\cite{axelrod2011domain,chinea2017adapting,zeng2019iterative,wang2020learning}, while \method focused on the training model which can complement with the data-driven methods perfectly. 

\subsection{Continual Learning}
The main idea of our approach is originated from the Continual Learning community \cite{parisi2019continual, kirkpatrick2017overcoming, mallya2018packnet, mallya2018piggyback, hung2019compacting, lee2019mixout}, as they all try to alleviate the \textit{catastrophic forgetting} problem.
\citet{mallya2018packnet, mallya2018piggyback, hung2019compacting} learn separate sub-networks for multiple tasks in computer vision, which inspires us with \method for machine translation domain adaptation.

\subsection{Model Pruning }
Our approach is also inspired by many studies of sparse networks \cite{frankle2018lottery, zhu2017prune, liu2018rethinking, masana2017domain}. 
\citet{frankle2018lottery, liu2018rethinking} reevaluate unstructured network pruning to highlight the importance of sparse network structure. 
\citet{zhu2017prune} introduce advanced pruning technique to compress the model.
\citet{sun2019learning} learn sparse sharing architecture for multi-task learning.
\citet{hung2019compacting} introduce compact parameter sub-network for continual learning. 
Different from these work, \method aims at finding the best sparse structure for a specific domain  based on an NMT model trained on large scale general domain data. Model pruning is an effective method for our approach.


\section{Conclusion and Future Work}
In this work, we propose \method, an effective way for adapting neural machine translation models which first generates an informative sub-network for the general domain via gradual pruning and then fine-tunes the unnecessary parameters for the target domain. By doing so, \method is able to retain as much general information as possible and alleviate the \textit{catastrophic forgetting} problems. Experiments show that the proposed \method outperforms fine-tuning and several strong baselines and it is shown to be much more robust compared to fine-tuning due to the complete retainment of the general information. Beyond that, \method can be extended to adapting multiple domains by iteratively pruning and tuning, which is naturally suitable for multi-lingual scenario. We leave the multi-lingual problem as our future work.

\section{Acknowledgements}
We would like to thank the anonymous reviewers for their valuable comments. We would also like to thank Zehui Lin, Yang Wei, Danqing Wang, 
Zewei Sun, and Kexin Huang for their useful suggestions and help with experiments. This work was supported by the National Key Research and Development Program of China (No. 2017YFB1002104).

\bibliography{aaai21}

\end{document}